%% file: main.tex
\documentclass{IEEEoj}
\usepackage{cite}
\usepackage{amsmath,amssymb,amsfonts}
\usepackage{algorithmic}
\usepackage{graphicx,color}
\usepackage{textcomp}
\usepackage{amsmath,amsfonts}
\usepackage{algorithmic}
\usepackage{algorithm}
\usepackage{array}
\usepackage[caption=false,font=normalsize,labelfont=sf,textfont=sf]{subfig}
\usepackage{textcomp}
\usepackage{stfloats}
\usepackage{url}
\let\labelindent\relax
\usepackage{enumitem}
\usepackage{verbatim}
\usepackage{graphicx}
\usepackage{cite}
\usepackage{xcolor}
\usepackage{booktabs}
\usepackage{multirow}
\usepackage{subfig}

\def\BibTeX{{\rm B\kern-.05em{\sc i\kern-.025em b}\kern-.08em
    T\kern-.1667em\lower.7ex\hbox{E}\kern-.125emX}}
\begin{document}


\title{Adapt under Attack and Domain Shift: Unified Adversarial Meta-Learning and Domain Adaptation for Robust Automatic Modulation Classification}

\author{Ali Owfi \IEEEauthorrefmark{1}, Amirmohammad Bamdad, Tolunay Seyfi \IEEEauthorrefmark{1}, Fatemeh Afghah \IEEEauthorrefmark{1}}
\affil{Department of Electrical and Computer Engineering, Clemson University}

\corresp{CORRESPONDING AUTHOR: Fatemeh Afghah (e-mail: fafghah@clemson.edu).}
\authornote{This material is based upon work supported by the National Science Foundation under Grant Numbers CNS-2202972, CNS- 2318726, and CNS2232048.}
\markboth{Adapt under Attack and Domain Shift: Unified Adversarial Meta-Learning and Domain Adaptation for Robust Automatic Modulation Classification}{Author \textit{Owfi et al.}}

\begin{abstract}

Deep learning has emerged as a leading approach for Automatic Modulation Classification (AMC), demonstrating superior performance over traditional methods. However, vulnerability to adversarial attacks and susceptibility to data distribution shifts hinder their practical deployment in real-world, dynamic environments. To address these threats, we propose a novel, unified framework that integrates meta-learning with domain adaptation, making AMC systems resistant to both adversarial attacks and environmental changes.
Our framework utilizes a two-phase strategy. First, in an offline phase, we employ a meta-learning approach to train the model on clean and adversarially perturbed samples from a single source domain. This method enables the model to generalize its defense, making it resistant to a combination of previously unseen attacks. Subsequently, in the online phase, we apply domain adaptation to align the model's features with a new target domain, allowing it to adapt without requiring substantial labeled data. As a result, our framework achieves a significant improvement in modulation classification accuracy against these combined threats, offering a critical solution to the deployment and operational challenges of modern AMC systems.
\end{abstract}


\maketitle

\input{1_intro}
\input{2_relatedwork}

\input{3_Preliminaries}
\input{4_problemFormulation}

\input{5_methodology}
\input{6_results}
\input{7_conclusion}

\bibliographystyle{IEEEtran}
\bibliography{main}

\end{document}

%% file: 1_intro.tex
\section{Introduction}

\IEEEPARstart {A}{utomatic} Modulation Classification (AMC) is a fundamental technology for enabling intelligent and adaptive wireless communication systems \cite{o2016radio}. Deep Learning (DL) has emerged as the leading approach for AMC, with numerous studies demonstrating its superior performance over traditional methods in classifying a wide range of modulation formats under various channel conditions \cite{o2016convolutional,rajendran2018deep,owfi2023meta}.

Despite these promising results \cite{o2017introduction}, the practical deployment of DL-based AMC models has been hindered by a critical vulnerability that affects the broader field of deep learning: susceptibility to adversarial attacks \cite{szegedy2013intriguing, zhang2024hyperadv}. These attacks involve the injection of low power perturbations into the input signal that, while often imperceptible, are designed to cause misclassification by the target model \cite{goodfellow2014explainingFSGM}. Numerous studies have explored this critical vulnerability of DL-based AMC models, addressing how the attacks decrease the accuracy of models \cite{sadeghi2018adversarial, lin2020threats, flowers2019evaluating}. Research has shown that such adversarial interference can be significantly more effective in degrading AMC performance than equivalent power additive white Gaussian noise (AWGN) \cite{maroto2022safeamc}.

To deal with this problem, various defensive mechanisms have been proposed, such as randomized smoothing and data augmentation \cite{kim2020vulnerability}, ensemble learning \cite{sahay2021deep}, and misclassification correction \cite{sahay2021deep}, although adversarial training, a process of augmenting the training dataset with adversarial examples, has been the most widely adopted defense mechanism \cite{liu2020adversarial, zhang2021adversarial}. However, while this technique improves robustness to known attack types \cite{sahay2021robust, maroto2022safeamc}, it often fails to generalize to unseen adversarial perturbations that the model will encounter when deployed in an online environment.

Our prior work \cite{ours} addressed this limitation by introducing a meta learning based adversarial training framework that explicitly optimized the AMC model for generalization to unseen black box attacks. In this framework, the model was trained across a distribution of adversarial perturbations to learn a robust initialization that could be quickly adapted to new attacks using only a few samples. This approach demonstrated strong generalization and adaptability under previously unobserved threats, even in zero shot and few shot scenarios.

However, our earlier framework assumed that the underlying data distribution in both offline and online phases remained consistent, meaning the training and deployment environments followed the same channel, hardware, and noise characteristics. This stationary setting is not realistic in practical wireless systems. This phenomenon, known as data distribution shift, causes a significant degradation in model performance when the model is deployed in a new environment (the target domain) that differs from its initial training environment (the source domain)\cite{sayyed2025resilient}. Assuming such shifts occur is essential to accurately reflect real world wireless systems, which experience frequent changes in signal-to-noise ratio (SNR), fading, interference profiles, and spectral conditions.

The challenge of distribution shift has been addressed in other areas of wireless communications. For example, domain adaptation techniques have been applied to maintain the performance of Radio Frequency (RF) fingerprinting systems under varying channel conditions \cite{zhang2021domain, restuccia2020deep}. Similarly, domain adaptation has been used for cross domain wireless signal recognition to overcome performance degradation due to environmental changes \cite{wang2020cross}. However, the challenge of handling both adversarial attacks and distribution shift has not been addressed for AMC. A deployed AMC model is not only a target for various forms of interference but also must operate in a continuously evolving signal environment.

To address this gap, in this paper we extend our previous meta learning framework by incorporating mechanisms for domain adaptation. Specifically, we remove the assumption of stationarity and propose a two-phase training methodology that enables the AMC model to learn feature representations that are simultaneously robust to adversarial perturbations and generalizable across domains.

We take a comprehensive approach to jointly address the challenges of adversarial robustness and domain generalization in practical, nonstationary wireless environments for AMC. Our framework unifies meta-learning, adversarial training, and domain adaptation into a single cohesive training and inference strategy, enabling AMC models to remain reliable under both adversarial attacks and data distribution shifts. To the best of our knowledge, this is the first AMC framework explicitly designed to tackle both issues simultaneously. Our main contributions are summarized as follows:

\begin{itemize}
    \item \textbf{Unified Robust AMC Framework:} We propose the first AMC framework that explicitly addresses two major practical challenges: domain shift caused by nonstationary wireless environments and unknown adversarial attacks during deployment.
    
    \item \textbf{Offline Phase Adversarial Training Framework:} Our meta-learning-based adversarial training strategy in the offline phase explicitly optimizes generalization across a distribution of attacks, resulting in robustness against unseen adversarial perturbations.

    \item \textbf{Online Phase Domain Adaptation Framework Using Pilot Labels:} We leverage a small number of labeled target-domain pilot signals to guide domain alignment during the online phase, maintaining performance under realistic domain shifts and label-scarce conditions.

    \item \textbf{Addressing Real-time AMC Limitations:} The proposed unified framework accounts for practical constraints often neglected in AMC, such as limited training time, scarce samples, and computational resources after deployment.

    \item \textbf{Significant Increased Efficiency:} Our framework substantially improves efficiency in terms of both sample usage and training time.
    
\end{itemize}

\begin{figure}{}
        \centering
        \includegraphics[width=0.9\columnwidth]{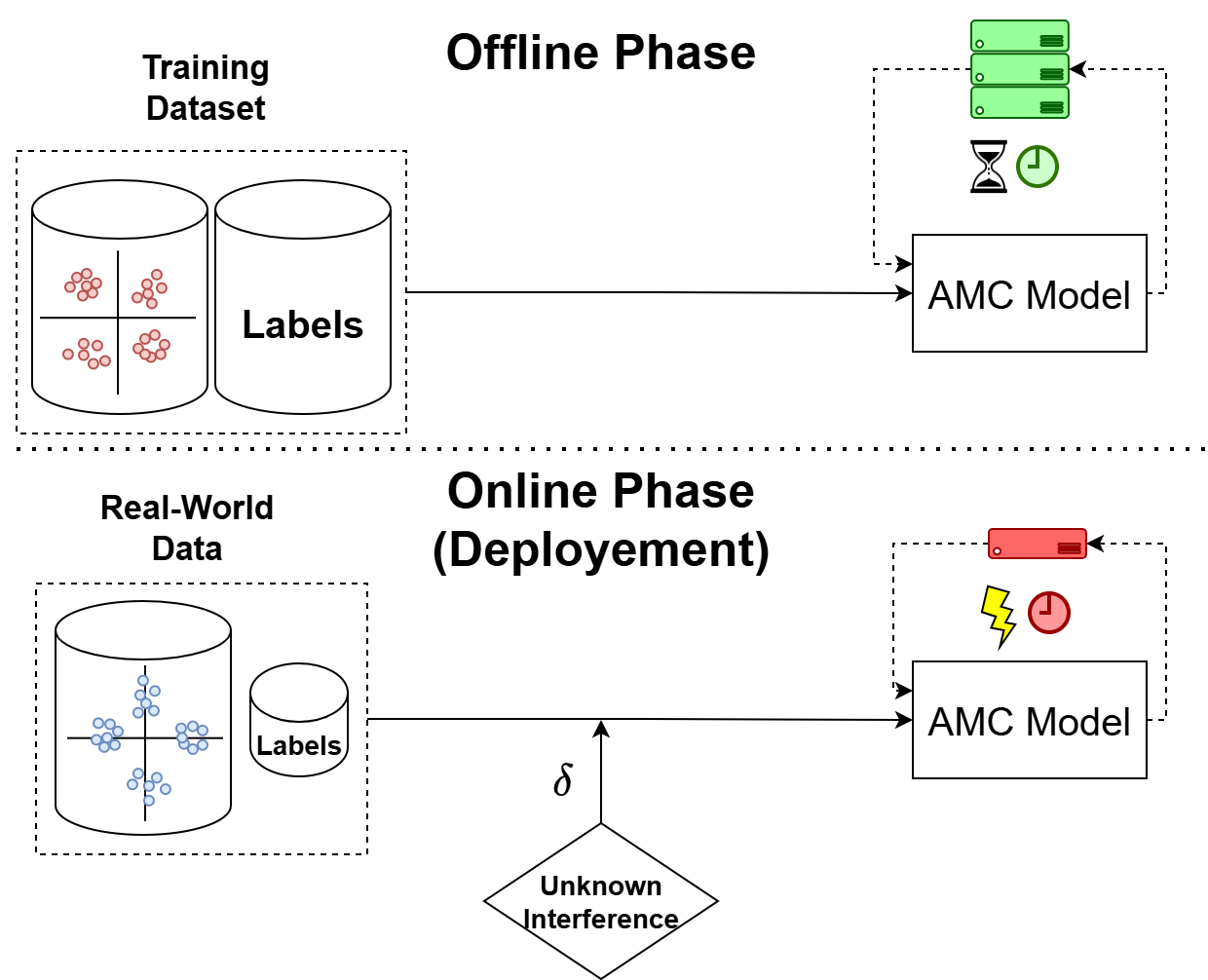} 
        \caption{Differences between the offline phase and the online deployment phase for AMC models. The underlying distribution between the training dataset used in the offline phase and the real-world data encountered during deployment are not the same. During deployment, only few labeled samples are available through pilot signals (if at all), and there are limitations on training time and resources. Lastly, unknown adversarial attacks can be encountered during the online phase.}
        \label{fig:AMC-offline-online}
        \vspace{-15pt}
 \end{figure}

%% file: 2_relatedwork.tex
\section{Related Work}

\subsection{Adversarial Attacks and Defenses in AMC}

The same complexity that enables deep learning (DL) models to perform well in AMC also makes them vulnerable to adversarial examples. This vulnerability was first discovered in the computer vision domain \cite{szegedy2013intriguing} and later shown to be a critical issue in wireless communications \cite{sadeghi2018adversarial, kim2020vulnerability}. An attacker can introduce imperceptible but malicious perturbations to radio signals, significantly degrading the performance of AMC models. Multiple attack mechanisms have been studied. Gradient-based attacks such as the Fast Gradient Sign Method (FGSM) \cite{goodfellow2014explainingFSGM}, Projected Gradient Descent (PGD) \cite{mkadry2017towardsPGD}, and the Momentum Iterative Method (MIM) \cite{dong2018boostingMIM} perturb the input along the direction of the loss gradient. More sophisticated optimization-based methods such as the Carlini and Wagner (C\&W) attack \cite{carlini2017towardsCW} craft minimal, high-impact perturbations. The practical use of these mechanisms is dictated by the attack scenario, which is defined by the attacker's level of knowledge about the model. In a white-box scenario, the attacker has full access to the model's parameters and gradients, allowing for the direct application of gradient-based or optimization-based methods. In a black-box scenario, the attacker has no such internal access. This often requires them to train a substitute model to approximate the target model's behavior and generate perturbations. Sadeghi et al.\ \cite{sadeghi2018adversarial} demonstrated that AMC models are vulnerable in both scenarios, successfully disrupting performance with white-box methods and a Principal Component Analysis (PCA)-based black-box attack. These vulnerabilities have been further validated under realistic wireless conditions\cite{flowers2019evaluating}.

To defend against such attacks, several techniques have been proposed. Early strategies include preprocessing-based defenses such as input denoising, randomized smoothing, and data augmentation, which aim to mitigate adversarial influence without modifying the model architecture \cite{kim2021channel}. Ensemble-based methods train multiple classifiers and fuse their decisions to improve robustness against a broader range of perturbations \cite{sahay2021deep}. However, adversarial training has emerged as the most prominent defense strategy \cite{liu2020adversarial, zhang2021adversarial}. This approach explicitly trains the model on adversarial examples, enabling it to recognize and resist perturbations similar to those encountered during training.

Despite its effectiveness, adversarial training suffers from limited generalization. A model trained against one attack (e.g., FGSM) often remains vulnerable to others (e.g., C\&W), especially if the training data lacks diversity in attack types \cite{maroto2022safeamc}. Furthermore, many defense strategies assume stationary or matched training and test conditions, ignoring the reality of shifting channel distributions. This highlights the need for robustness mechanisms that are attack-agnostic and capable of adapting to both adversarial and non-stationary environments.

These limitations motivate the unified treatment of adversarial robustness and domain adaptation explored in this work, where the goal is to sustain classification performance under simultaneous perturbation and distribution shift.

\subsection{Domain Adaptation in Wireless Communications}

Recent efforts have increasingly emphasized the need for resilient automatic modulation classification (AMC) models that can generalize across diverse operational settings without retraining. This reflects a growing recognition that real-world wireless environments are rarely stationary or fully observable during training, necessitating mechanisms to bridge the distribution gap between source and deployment domains. In such conditions, even moderate mismatches in SNR or propagation characteristics can significantly degrade classification accuracy, underscoring the importance of domain adaptation techniques for practical AMC deployment. To address this, \cite{wang2020cross} propose a two-level data augmentation method that enhances generalization across mismatched channel and SNR conditions between source and target domains.

Expanding on this idea, \cite{wang2024sigda} introduce \textit{SigDA}, a superimposed domain adaptation framework that jointly aligns both feature and prediction distributions via dual-branch adaptation. SigDA integrates attention mechanisms to emphasize domain-invariant features and employs adversarial training to refine alignment. These techniques reflect broader trends in wireless domain adaptation, including adversarial alignment, supervised fine-tuning with pseudo-labeled target data, and discrepancy-based methods such as Maximum Mean Discrepancy (MMD).

To reduce the reliance on large labeled target datasets, \textit{PSRNet} \cite{xing2024psrnet} incorporates few-shot learning into the domain adaptation process. It introduces a paired-sample relation network that learns class-level similarity scores from only a few labeled target samples. By constructing cross-domain modulation pairs and training a relation-based classifier, PSRNet achieves robust generalization even under severe distribution shifts such as multipath fading and SNR mismatch.

In the fully unsupervised setting, \cite{deng2024cdamc} propose a multimodal progressive adaptation network that fuses amplitude, phase, and I/Q signal modalities. Their method incrementally refines pseudo-labels during training, allowing the model to extract domain-invariant yet class-discriminative features across varying channel and SNR conditions without access to target labels.

A complementary approach is presented in \cite{hou2023mdf}, which leverages multiple diverse source domains to proactively enhance generalization. Their multi-domain fusion strategy uses attention-based feature aggregation to combine knowledge from different wireless environments, reducing dependence on test-time adaptation and increasing resilience to unseen target distributions.

Finally, \cite{hassan2025mfuda} propose the M-FUDA framework for Wi-Fi CSI-based activity recognition. Although designed for a different wireless task, M-FUDA demonstrates key transferable concepts applicable to AMC under domain shift. It aligns temporal and spectral features across domains in a fully unsupervised manner, guided by class-wise feature centroids and adaptive margin constraints to preserve discriminative structure.

Together, these works mark significant progress toward robust signal classification under domain shift. However, they do not account for the simultaneous presence of adversarial interference, a realistic and critical challenge in open wireless environments \cite{hassan2025mfuda, hao2023automatic, deng2024cdamc, xing2024psrnet, wang2024sigda}. In contrast, our framework is the first to jointly address both adversarial robustness and domain adaptation, offering a unified approach to sustain AMC performance when facing both distributional and adversarial threats..

\subsection{Meta-Learning for Adaptation and Generalization}

Meta-learning has emerged as a promising solution for enhancing generalization in wireless signal classification tasks under limited supervision or distribution shifts. In the context of AMC, prior work has explored various formulations of meta-learning tailored to noisy training conditions, few-shot inference, and adaptation across non-stationary environments.

A prominent line of work targets label imperfection and supervision noise. To address robustness under such conditions, \cite{hao2024meta} propose a teacher-student hybrid network (TSHN) that combines meta-learning with label noise distillation. The teacher is trained on a small set of trusted clean samples using a meta-objective, while the student is guided through noisy examples. Their design incorporates multiple signal modalities—including I/Q, amplitude, and phase—and fuses them through a multimodal encoder to improve robustness against corrupted training data.

Complementary efforts focus on few-shot and open-set generalization. \cite{zhang2024fsos} introduce FSOS-AMC, a meta-learning framework for few-shot modulation classification in the presence of previously unseen classes. Their approach leverages a multi-scale attention network to extract hierarchical features and uses meta-learned prototypes to guide inference. An auxiliary module performs open-set rejection to manage unknown modulation types encountered at test time.

Another direction explores meta-initialization for efficient adaptation across dynamic wireless environments. \cite{park2019demod} apply gradient-based meta-learning techniques, including MAML, FOMAML, and REPTILE, to enable demodulators to rapidly adapt to new channel conditions using only a few pilot transmissions. Although not directly designed for AMC, this formulation highlights the potential of meta-initialization in non-stationary signal processing tasks where rapid task adaptation is essential.

While these approaches demonstrate the flexibility of meta-learning in handling various challenges in AMC—such as supervision noise, limited data, and open-class generalization—they typically address only one axis of difficulty. In contrast, our work jointly considers both adversarial robustness and domain adaptation within a unified meta-learning framework that operates under real-world wireless constraints.

%% file: 3_Preliminaries.tex
\section{Preliminaries} \label{preliminaries}

\subsection{Fundamentals of Adversarial Attacks in Deep Learning-Based AMC}

A great success of deep learning models is their ability to learn complex, high-dimensional decision boundaries directly from data. However, a critical vulnerability arises from the very nature of these boundaries. While these models perform well on standard data, the decision boundaries they learn are fragile. This means that a small, carefully crafted perturbation, imperceptible to humans, is sufficient to push an input sample across a decision boundary, causing the model to produce an incorrect prediction. This perturbed sample is known as an adversarial example, and the process of creating it is an adversarial attack.
In the context of AMC, an input signal x with true modulation class \(y \) is fed to a DL-based classifier, denoted by the function \(f \) with parameters $\theta$. An adversary's goal is to generate a small perturbation, $\delta$, such that when added to the original signal, the resulting adversarial signal is misclassified. The perturbation $\delta$ is typically constrained by an $l_{p}$ norm, and in the context of wireless communication, $l_{2}$ norm is the most common norm to be used. Generating this perturbation can be formed as a constrained optimization problem. The adversary’s objective is to maximize the loss function \(L \):

\begin{equation}
\begin{aligned}
& \max_{\delta} \mathcal{L}(\theta, f(x + \delta), y) 
\\
&\text{s.t.} \quad \|\delta\|_p \leq \epsilon \quad
\label{eq:adv_attack}
\end{aligned}
\end{equation}
 
Various methods have been developed to find effective, approximate solutions for this problem. Below, we outline the attack methods that are used in this paper:

\begin{itemize}[leftmargin=*]

    \item \textbf{Fast Gradient Sign Method (FGSM) Attack:} FGSM \cite{goodfellow2014explainingFSGM} is a fast, single-step attack. It generates a perturbation that increases the loss. The perturbation is calculated by taking the sign of the gradient of the loss function with respect to the input. This direction is an efficient way to increase the model's loss in a single step. The adversarial example is generated as: 
\begin{equation}
\begin{aligned}
\text{x}_{adv} = \text{x} + \epsilon \cdot \text{sign}(\nabla_x \mathcal{L}(\theta,x, y))
\label{eq:FSGM}
\end{aligned}
\end{equation}
where $\epsilon$ is a small scalar controlling the magnitude of the perturbation.

\item \textbf{Projected Gradient Descent (PGD) Attack:}
 PGD \cite{mkadry2017towardsPGD,kurakin2016adversarialPGD} is an iterative version of FGSM. It takes multiple small steps in the direction of the gradient, projecting the result onto the allowed perturbation region after each step. The update rule for each step is :
 \begin{equation}
 x^{t+1} = \Pi _{\epsilon} \left( x^t + \alpha \cdot \text{sign}(\nabla_x \mathcal{L}(\theta,x, y)) \right)
 \end{equation}
where $\alpha$ is the step size and $\Pi_{\epsilon}(.)$ is the projection function that clips the total perturbation.

\item\textbf{Momentum Iterative Method (MIM) Attack:}

MIM \cite{dong2018boostingMIM} is another enhanced iterative attack like PGD but using momentum. Standard iterative methods can get stuck in poor local optima; however, MIM uses momentum to stabilize the update steps and escape the local maxima. The update at each step is:
\begin{equation}
\begin{aligned}
g^{t+1} &= \mu \cdot g^t + \frac{\nabla_x \mathcal{L}(\theta, x, y)}{\|\nabla_x \mathcal{L}(\theta, x, y)\|_1} \\
x^{t+1} &= x^t + \alpha \cdot \text{sign}(g^{t+1})
\label{eq:MIM}
\end{aligned}
\end{equation}
 where $\mu$ is the momentum coefficient.

\item\textbf{Carlini and Wagner (C\&W) Attack:}
 The Carlini and Wagner (C\&W) attack \cite{carlini2017towardsCW} is a powerful, optimization-based attack that, instead of trying to maximize the classification loss, searches for the minimum perturbation $\delta$ that causes a misclassification to a specific target class \(t \)  (or any class other than the true one). Formulated as an optimization problem, the objective function of it is:  

 \begin{equation}
 \begin{aligned}
 & \min_{\delta} \|\delta\|_2^2 + c \cdot \mathcal{L}(f(x + \delta), y) \\
 & \text{s.t.} \quad x + \delta \in [0, 1]^n
 \end{aligned}
 \end{equation}
 
The attack aims to make the logit for a false class larger than the logit for the true class (maximizing their distance) while minimizing the perturbation.

\item\textbf{Principal Component Analysis (PCA) Attack:}
This method \cite{sadeghi2018adversarial} aims to find a single, universal perturbation vector \(v \) that can be added to any input signal from a given data distribution to cause misclassification. This attack uses PCA to identify the principal direction of example data variance. The hypothesis is that perturbations along these directions are more likely to push samples across the model's decision boundary.

 \end{itemize}

\subsection{Meta-Learning Principles for Rapid Generalization and Robustness}

Meta-learning, often referred to as “learning to learn,” is a framework designed to enable models to adapt quickly to new tasks using only a few training examples. Rather than training a model to perform well on a single task, meta-learning optimizes the learning algorithm itself across a distribution of tasks. This higher-order optimization enables the model to internalize an inductive bias that facilitates rapid generalization, even in scenarios with limited or shifted data—conditions commonly encountered in real-world wireless environments such as AMC.

Formally, consider a distribution over tasks \(\mathcal{T} \sim p(\mathcal{T})\), where each task consists of its own dataset split into a support set which is used for adaptation and a query set which is used for evaluation. The meta-learning algorithm seeks a parameter initialization \(\theta\) such that, for any new task \(\mathcal{T}_i\), a small number of gradient updates on the support set leads to strong generalization on the query set.

Several algorithmic strategies have emerged to realize this goal: Model-Agnostic Meta-Learning (MAML), First-Order MAML (FOMAML), and Reptile.

\begin{itemize}[leftmargin=*]

    \item \textbf{Model-Agnostic Meta-Learning (MAML):}
    MAML \cite{finn2017model} formulates meta-learning as a bi-level optimization problem. During meta-training, it learns an initialization \(\theta\) that can be quickly adapted via a few gradient steps to perform well on new tasks. The update rule involves an inner loop for task-specific adaptation and an outer loop for meta-optimization:
    \begin{equation}
    \theta \leftarrow \theta - \beta \nabla_{\theta} \sum_{\mathcal{T}_i \sim p(\mathcal{T})} \mathcal{L}_{\mathcal{T}_i}(f_{\theta'_i}),
    \end{equation}
    where \(\theta'_i = \theta - \alpha \nabla_{\theta} \mathcal{L}_{\mathcal{T}_i}(f_{\theta})\) is the adapted parameter set for task \(\mathcal{T}_i\), and \(\alpha, \beta\) denote the inner and outer learning rates, respectively. The second-order gradient computation enables precise tracking of how adaptation influences performance, but incurs higher computational cost.

    \item \textbf{First-Order MAML (FOMAML):}
    FOMAML \cite{nichol2018reptile} simplifies MAML by omitting the second-order derivative term in the meta-gradient. This approximation dramatically reduces computational overhead while preserving much of MAML’s adaptability. The meta-update is:
    \begin{equation}
    \theta \leftarrow \theta - \beta \sum_{\mathcal{T}_i \sim p(\mathcal{T})} \nabla_{\theta'} \mathcal{L}_{\mathcal{T}_i}(f_{\theta'_i}),
    \end{equation}
    where \(\theta'_i\) is obtained via a standard gradient descent step as in MAML. Despite the simplification, FOMAML has been shown to achieve comparable performance in many meta-learning benchmarks, including signal classification tasks.

    \item \textbf{Reptile:}
    Reptile \cite{nichol2018reptile} takes a more heuristic approach to meta-learning by forgoing explicit gradient tracking. It performs multiple steps of task-specific training and then moves the initialization \(\theta\) toward the adapted parameters. For a given task \(\mathcal{T}_i\), the update rule is:
    \begin{equation}
    \theta \leftarrow \theta + \epsilon (\theta'_i - \theta),
    \end{equation}
    where \(\theta'_i\) results from several steps of SGD on task \(\mathcal{T}_i\). Reptile is computationally efficient and well-suited for non-convex models, making it appealing for resource-constrained wireless learning systems.

\end{itemize}

\subsection{Domain Adaptation Under Covariate Shift: Concepts and Mechanisms}

Deep learning models typically assume that training and test data are drawn from the same underlying distribution. However, in real-world wireless scenarios, especially in spectrum sensing or modulation classification tasks, this assumption is often violated due to variations in channel conditions, hardware impairments, or deployment environments. This mismatch, known as domain shift, degrades the generalization performance of models trained solely on data from a fixed source distribution.

Domain adaptation (DA) aims to bridge this gap by transferring knowledge from a labeled source domain \(\mathcal{D}_S = \{(x_i^S, y_i^S)\}_{i=1}^{N_S}\) to an unlabeled or sparsely labeled target domain \(\mathcal{D}_T = \{x_j^T\}_{j=1}^{N_T}\), where the marginal distributions differ, i.e., \(P_S(x) \neq P_T(x)\), but the conditional distributions are assumed to be shared, \(P_S(y|x) \approx P_T(y|x)\). This setting is referred to as covariate shift, and it poses unique challenges in signal classification where training data may be collected in controlled lab conditions, but deployment occurs in more complex field environments.

To overcome this challenge, a variety of domain adaptation methods have been proposed. In this section, we highlight one of the most widely-used frameworks: Domain-Adversarial Neural Networks (DANN).

\begin{itemize}[leftmargin=*]

    \item \textbf{Domain-Adversarial Neural Network (DANN):}
    DANN \cite{ganin2016domain} is a gradient-based adversarial method for unsupervised domain adaptation that introduces a domain discriminator alongside the feature extractor and classifier. The core idea is to learn a feature representation \(f(x)\) that is both discriminative for the source task and domain-invariant across source and target domains.

    The architecture includes three components:
    \begin{enumerate}
        \item A feature extractor \(G_f(x; \theta_f)\)
        \item A label predictor \(G_y(G_f(x); \theta_y)\) trained on source labels
        \item A domain classifier \(G_d(G_f(x); \theta_d)\) trained to distinguish source vs. target domains
    \end{enumerate}

    The feature extractor is optimized to fool the domain classifier via a gradient reversal layer (GRL), effectively reversing gradients during backpropagation. The resulting optimization objective is:
    \begin{equation}
    \min_{\theta_f, \theta_y} \max_{\theta_d} \ \mathcal{L}_{\text{task}}(\theta_f, \theta_y) - \lambda \mathcal{L}_{\text{domain}}(\theta_f, \theta_d)
    \end{equation}

    where \(\mathcal{L}_{\text{task}}\) is the classification loss on source labels and \(\mathcal{L}_{\text{domain}}\) is the domain classification loss over both source and target samples. The hyperparameter \(\lambda\) balances the two objectives.

    Through adversarial training, DANN encourages the learned features to be both predictive for the task and indistinguishable across domains, thus enabling better transfer performance under covariate shift.

\end{itemize}

%% file: 4_problemFormulation.tex
\section{Problem Formulation}

\subsection{Learning Under Distribution Shift and Adversarial Conditions}

We consider a DL-based AMC model at the receiver side, denoted by $f_{\theta}(.): \mathbb{R}^{2\times\lambda} \rightarrow \mathbb{R}^{C}$. The AMC model predicts one of the $C$ possible modulations when given an IQ signal shaped as $2\times\lambda$ as the input. We consider two separate phases as shown in figure \ref{fig:AMC-offline-online}, the offline training phase and the online real-world deployment phase.

During the offline phase, the AMC model has access to a sufficiently large labeled training dataset denoted by $\mathcal{D}_s = \left\{ \left( x^s_i, y^s_i \right) \right\}_{i=1}^{N_s}$. $\mathcal{D}_s$ is drawn from a source domain $\mathcal{D}_{\text{source}} = \{\mathcal{X}, P_{\text{source}}(x, y)\}$, where $P_{\text{source}}(x, y)$ represents the joint probability distribution over input signals $x \in \mathcal{X}$ and their corresponding modulation labels $y$.

After the offline training is done, the AMC model is deployed to a real-time online system. During the online phase, the AMC model faces multiple challenges:

\begin{itemize}[leftmargin=*]
    \item \textbf{Domain shift:} Due to the dynamic nature of wireless communication systems, the received signals during the online phase — denoted by $\mathcal{D}_t = \left\{ \left( x^t_i, y^t_i \right) \right\}_{i=1}^{N_t}$, where $t$ indicates the target domain — will not follow the same distribution as the offline training data $\mathcal{D}_s$. More formally, $P_{\text{source}}(x, y) \neq P_{\text{target}}(x, y)$, where $\mathcal{D}_s$ and $\mathcal{D}_t$ are drawn from a source domain $\mathcal{D}_{\text{source}} = \left\{ \mathcal{X}, P_{\text{source}}(x, y) \right\}$ and a target domain $\mathcal{D}_{\text{target}} = \left\{ \mathcal{X}, P_{\text{target}}(x, y) \right\}$, respectively.
    \item \textbf{Semi-supervised/unsupervised instead of supervised:} Unlike in the offline phase, the AMC model will not have access to the labels for most received signals during the online phase, if any at all. In a realistic scenario, the only few labeled samples available to the AMC model would be obtained through pilot signals. Formally, the set of received signals in the online phase consists of a labeled and an unlabeled part, denoted as $\mathcal{D}_t = \mathcal{D}_t^l \cup \mathcal{D}_t^u$, where $\mathcal{D}_t^l = \left\{ \left( x^t_i, y^t_i \right) \right\}_{i=1}^{N_t^l}$ represents the small set of labeled samples, and $\mathcal{D}_t^u = \left\{ x^t_j \right\}_{j=1}^{N_t^u}$ represents the large set of unlabeled samples, with $|\mathcal{D}_t^l| \ll |\mathcal{D}_t^u|$.
    \item \textbf{Interference from black-box adversarial attack:} As shown in Figure \ref{fig:adversarial_interference}, we consider the presence of an interferer employing a black-box adversarial attack, injecting adversarial perturbations into the transmitted signals to degrade the model's classification performance without direct access to the model parameters. Formally a received signal $x^t_i$ can be defined as $x^t_i = x^{t, \text{clean}}_i + \delta. + n$ where $x^{t, \text{clean}}_i$ is the clean transmitted signal, $\delta$ is the injected adversarial perturbation, and $n$ is the added noise from the channel.
\end{itemize}

To summarize, in the online phase, the AMC model has to deal with the domain shift as well as the adversarial attack from the interferer. To do this, the AMC model can only access the offline training source domain data $\mathcal{D}_s = \left\{ \left( x^s_i, y^s_i \right) \right\}_{i=1}^{N_s}$, the unlabeled target domain data $\mathcal{D}_t^u = \left\{ x^t_j \right\}_{j=1}^{N_t^u}$ in addition to the small set of labeled samples from the target domain $\mathcal{D}_t^l = \left\{ \left( x^t_i, y^t_i \right) \right\}_{i=1}^{N_t^l}$ only if the pilot signals are available. The performance of the AMC model is only assessed by its adaptation capabilities during the online phase to the target domain and not based on its performance during the offline phase.



\subsection{Adversarial Attack Interference}

\begin{figure}{}
        \centering
        \includegraphics[width=0.7\columnwidth]{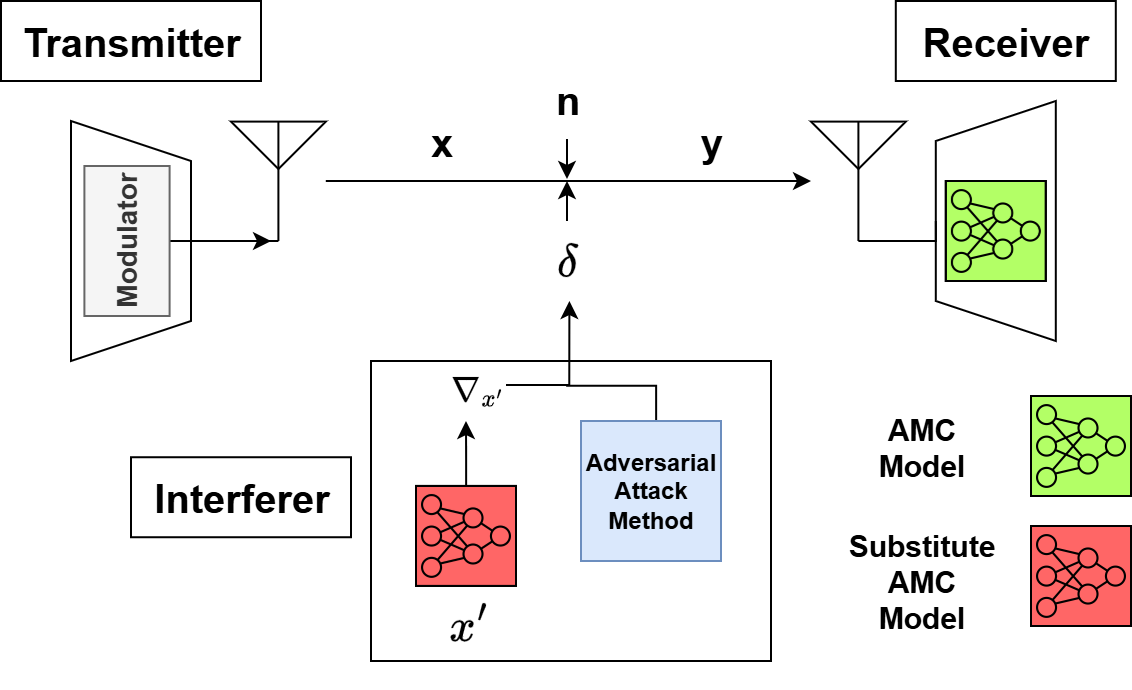} 
        \caption{Black-box adversarial attack on an AMC model in a wireless communication system. The interferer generates adversarial perturbations using a substitute model and injects them into the transmitted signal to deliberately induce misclassification at the DL-based AMC model incorporated into the receiver. The perturbation is unknown to the AMC model.}
        \label{fig:adversarial_interference}
 \end{figure}
 
We consider a wireless communication scenario in which a source of interference deliberately targets the DL-based AMC model deployed at the receiver using an unknown adversarial attack, as illustrated in Fig.~\ref{fig:adversarial_interference}. Unlike conventional interference, such adversarial attacks introduce carefully crafted, minimal-power perturbations designed to induce maximal degradation in the decision performance of the DL-based AMC model at the receiver.

Ideally, an adversary would require access to the target AMC model’s parameters, architecture, or prediction outputs to compute effective perturbations. However, in practical wireless environments, such access is rarely feasible. Consequently, the adversary resorts to training a substitute neural network that approximates the behavior of the target model based on observed input-output pairs. This surrogate model is then used to generate adversarial perturbations via gradient-based methods.

As the adversary lacks direct access to the target model’s internal information, this constitutes a black-box adversarial attack. The perturbation is computed solely with respect to the substitute model but is transferred to the target AMC model by exploiting the transferability property of adversarial examples\cite{szegedy2013intriguing}.

Formally, given a clean dataset $\mathcal{D} = \{(\mathbf{x}_i, y_i)\}_{i=1}^N$, perturbations are computed as
\begin{equation}
    \delta_i = \mathcal{A}(\mathbf{x}_i, f_{\text{sub}}, \epsilon)
\end{equation}
where $\mathcal{A}$ is an adversarial attack method, $f_{\text{sub}}$ is the substitute model, and $\epsilon$ controls the perturbation strength. The adversarial example is then $\mathbf{x}^{\text{adv}}_i = \mathbf{x}_i + \delta_i$.

Importantly, the structure of such perturbations is inherently unknown to the receiver and the AMC model. Given the unbounded space of possible substitute models and adversarial attack algorithms, the resulting interference is unpredictable and should be treated as an unknown attack from the perspective of the AMC model.

%% file: 5_methodology.tex
\section{Methodology} \label{method}

\subsection{Proposed Offline Training Method}

\begin{figure*}{}
        \centering
        \includegraphics[width=0.8\textwidth]{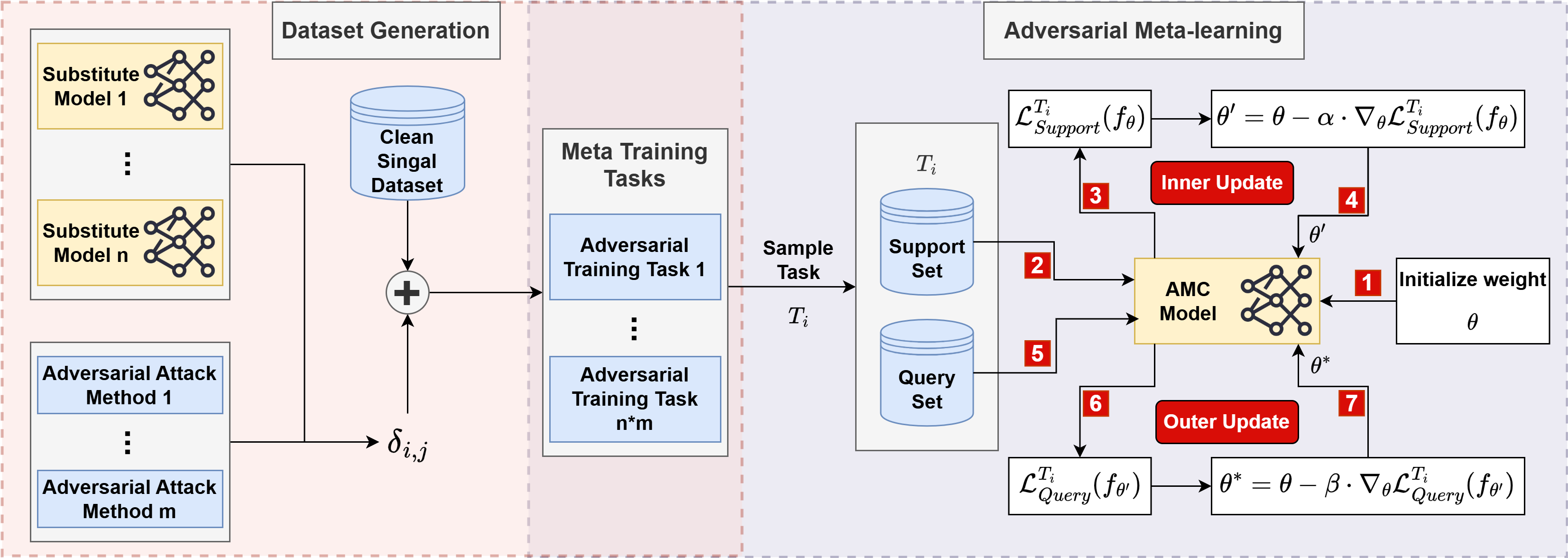} 
        \caption{Proposed meta-learning-based adversarial training framework for AMC models. The numbers in the adversarial meta-learning part denote the order of steps. The adversarial meta-learning section is depicted based on MAML, but theoretically, any model-agnostic meta-learning algorithm can be used as well. }
        \label{fig:adv_meta_amc}
        \vspace{-15pt}
 \end{figure*}

 \begin{algorithm}
\caption{Proposed offline training framework based on adversarial meta learning.}
\begin{algorithmic} 
\REQUIRE $\mathcal{A} = \{a_1, a_2, ..., a_n\}$ $\mathcal{S} = \{s_1, s_2, ..., s_m\}$, a set of adversarial attack methods and a set of substitute models 
\REQUIRE Clean signal dataset $\mathcal{D} = \{(x_k, y_k) \mid k = 1, \dots, N\}$ where $x_k$ is the received signal and $y_k$ is the modulation class

\STATE Randomly initialize weights $\theta$ for AMC model $f$ 
\FORALL{outer loop iterations}

\STATE Sample $a_i$ from $\mathcal{A}$ and $s_j$ from $\mathcal{S}$
\STATE Train $s_j$ using $\mathcal{D}$ 
\STATE Generate perturbation $\delta_{i,j}$ based on $a_i(s_j, \mathcal{D})$ 
\STATE $\mathcal{D}_{i,j} = \{(x_k+\delta_{i,j}, y_i) \mid k = 1, \dots, N\}$
\STATE Split $\mathcal{D}_{i,j}$ samples to $support$ and $query$ sets
\STATE Set $\theta' = \theta$

\FORALL{inner loop iterations}
\STATE Using $D_{i,j}$'s $support$ set compute loss $\mathcal{L}^{D_{i,j}}_{support}(f_{\theta'})$
\STATE Update $\theta^{'} \leftarrow \theta^{'} - \alpha\nabla_{\theta^{'}}\mathcal{L}^{D_{i,j}}_{support}(f_{\theta'})$
\ENDFOR

\STATE Using $D_{i,j}$'s $query$ set compute loss  $\mathcal{L}^{D_{i,j}}_{query}(f_{\theta'})$

\STATE Update $\theta \leftarrow \theta-\beta\nabla_{\theta}\mathcal{L}^{D_{i,j}}_{query}(f_{\theta'})$

\ENDFOR

\end{algorithmic}
\label{alg:META-adv}
\end{algorithm}

\subsubsection{Adversarial Training}

To improve resilience against adversarial attacks, a natural defense is to incorporate adversarial examples directly into the training process. By exposing the model to perturbed inputs during training, the classifier’s decision boundaries become more robust to intentionally crafted perturbations. This process, known as \textit{adversarial training}, modifies the standard training loop by generating adversarial examples for each batch and updating model parameters accordingly.

Adversarial examples are generated using a \textit{substitute model}, which approximates the target model’s decision boundaries. This mimics a realistic adversary that crafts perturbations based on a surrogate model. Integrating these examples into training helps the AMC model develop robustness against such attacks.

In each training iteration, a batch of clean samples $\{(\mathbf{x}_i, y_i)\}_{i=1}^N$ is drawn, adversarial examples are generated, and the AMC model $f_\theta$ is trained to minimize the adversarial loss:
\begin{equation}
    \mathcal{L}_{\text{adv}} = \frac{1}{N} \sum_{i=1}^N \mathcal{L}(f_\theta(\mathbf{x}^{\text{adv}}_i), y_i)
\end{equation}
The adversarial loss is then combined with the loss calculated from the clean data for better balance and robustness:
\begin{equation}
    \mathcal{L}_{\text{total}} = \lambda \cdot \frac{1}{N} \sum_{i=1}^N \mathcal{L}(f_\theta(\mathbf{x}_i), y_i) + (1 - \lambda) \cdot \mathcal{L}_{\text{adv}}
\end{equation}
where $\lambda \in [0, 1]$ controls the trade-off.

\subsubsection{Meta- Learning-based Adversarial Training}

While adversarial training improves robustness against known attacks, its generalization to unseen or adaptive perturbations remains limited, as it is constrained by the finite set of attack strategies used during training. To overcome this limitation, we adopt a meta-learning-based adversarial training framework for AMC models, enabling the model to learn transferable adversarial robustness strategies from a diverse set of adversarial tasks in the offline phase and efficiently generalize to new, unseen attacks in the online phase.

In this framework, each \textit{meta-training task} corresponds to an adversarial training scenario generated by a unique combination of an adversarial attack method and a substitute model. This task construction strategy exposes the AMC model to multiple attack behaviors and perturbation patterns during meta-training, thereby approximating the diversity of potential adversarial scenarios the model may face in deployment.

Formally, let \(\mathcal{T} = \{ T_1, T_2, \dots, T_M \}\) denote the set of meta-training tasks. Each task \(T_i\) is defined by generating adversarial perturbations using a specific adversarial attack method \(\mathcal{A}_i\) applied to a substitute model \(f_{\text{sub},i}\) with parameters \(\theta_{\text{sub},i}\). The adversarial perturbations \(\delta^{(i)}\) for task \(T_i\) are computed as:
\begin{equation}
    \delta^{(i)} = \mathcal{A}_i(\mathbf{x}, y, f_{\text{sub},i}, \epsilon_i)
\end{equation}
The adversarial training dataset for task \(T_i\) is then constructed by applying \(\delta^{(i)}\) to a clean dataset.

We employ the \textit{Model-Agnostic Meta-Learning (MAML)} algorithm \cite{finn2017model} to optimize the AMC model parameters for rapid adaptation to new adversarial attack scenarios. The meta-learning procedure consists of an \textit{inner loop} and an \textit{outer loop}.

In each outer-loop iteration:
\begin{itemize}
    \item A batch of meta-training tasks is sampled from \(\mathcal{T}\).
    \item For each task \(T_i\), the AMC model \(f_\theta\) is first adapted to task-specific parameters \(\theta'_i\) by performing one or more gradient descent updates on the task's \textit{support set}:
    \begin{equation}
        \theta'_i = \theta - \alpha \nabla_\theta \mathcal{L}^{T_i}_{\text{support}}(f_\theta)
    \end{equation}
    \item The adapted model \(f_{\theta'_i}\) is then evaluated on the task's \textit{query set} to compute the query loss \(\mathcal{L}^{T_i}_{\text{query}}\).
    \item The initial parameters \(\theta\) are updated using the aggregated query losses across all tasks:
    \begin{equation}
        \theta \leftarrow \theta - \beta \nabla_\theta \sum_{i} \mathcal{L}^{T_i}_{\text{query}}(f_{\theta'_i})
    \end{equation}
\end{itemize}

After several outer-loop iterations, the resulting parameters \(\theta^*\) serve as a robust initialization that enables the AMC model to quickly adapt to new adversarial scenarios with minimal updates, or to maintain satisfactory robustness even without adaptation if no online samples are available.

The critical advantage of this meta-learning-based adversarial training framework is its ability to significantly improve the AMC model's robustness to unseen adversarial attacks by optimizing the model over a distribution of diverse adversarial training tasks. By learning how to quickly adapt to new perturbation patterns during meta-training, the model becomes capable of generalizing its robustness beyond the specific attack scenarios encountered in the offline phase. As a result, even with scarce or rapidly acquired adversarial samples in the online phase, resulting in a few-shot learning scenario, the AMC model can effectively adapt its decision boundaries to counter previously unseen adversrail attacks.

\subsection{Proposed Online Adaptation Method}

\begin{figure*}{}
        \centering
        \includegraphics[width=0.8\textwidth]{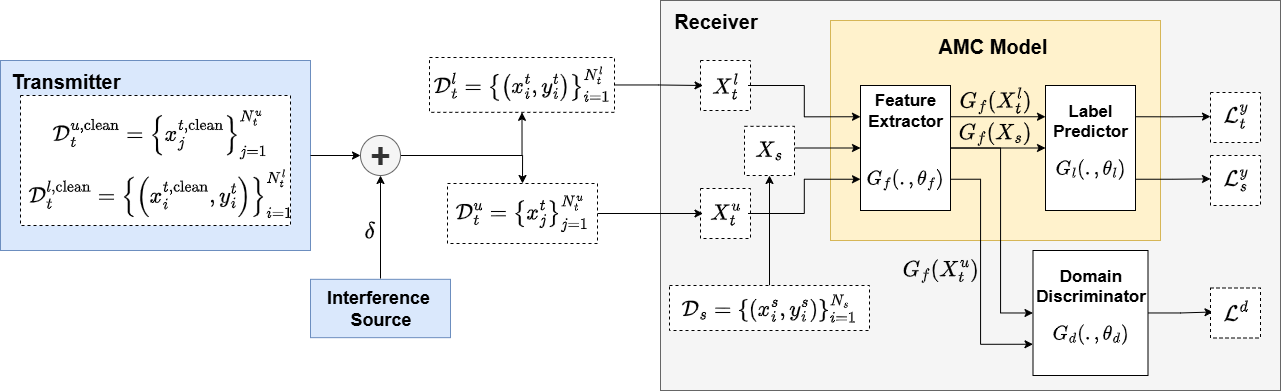} 
        \caption{Proposed online adaptation framework.}
        \label{fig:DA_online}
        \vspace{-15pt}
 \end{figure*}

In a realistic deployment scenario, the data distribution encountered during the online phase will differ from that of the offline training phase, resulting in a domain shift. Furthermore, full access to labeled online phase data is typically unavailable. The AMC model must therefore rely on a limited number of labeled samples alongside a large set of unlabeled samples to adapt and fine-tune itself to the changing conditions in real time. Formally, during the online phase the AMC model has access to the unlabeled target domain data $\mathcal{D}_t^u = \left\{ x^t_j \right\}_{j=1}^{N_t^u}$, and a small set of labeled target domain data $\mathcal{D}_t^l = \left\{ \left( x^t_i, y^t_i \right) \right\}_{i=1}^{N_t^l}$ from the online phase. Additionally the offline training source domain data $\mathcal{D}_s = \left\{ \left( x^s_i, y^s_i \right) \right\}_{i=1}^{N_s}$ is also available. 

The conventional way to fine-tune the AMC model to the target domain data would be perform a standard fine-tuning step on the the small labeled set $\mathcal{D}_t^l$. While this will be more effective than simply deploying the model that was only trained on the offline phase source data $\mathcal{D}_s$, it fails to utilize any information from the majority of target domain data $\mathcal{D}_t^u$ and the relation between the two related but different data domains, $\mathcal{D}_t$ and $\mathcal{D}_s$. Therefore, we propose the an online phase fine-tuning process based through the concept of domain adaption and based on the Domain-Adversarial Neural Network (DANN)\cite{ganin2016domain} method. 

The overall proposed framework can be seen in Figure \ref{fig:DA_online}. The left side of the figure depicts the process by which the online phase target domain data is created. A set of clean unlabeled $\mathcal{D}_{t}^{u,clean}$ and labeled $\mathcal{D}_{t}^{l,clean}$ signals are transmitted from the transmitter. The signals go through the channel, where they are added with noise and adversarial interference perturbations from the adversary, resulting in the received signals $\mathcal{D}_{t}^u$ and $\mathcal{D}_{t}^l$.

At the receiver, to perform domain adaptation, we need three components: a feature extractor \(G_f(x; \theta_f)\), a label predictor \(G_y(G_f(x); \theta_y)\), and a domain classifier \(G_d(G_f(x); \theta_d)\). \(G_f\) and \(G_y\) correspond to the main layers and the final prediction layer of the AMC model, respectively, which are pre-trained on source domain data during the offline phase. The domain discriminator \(G_d\), on the other hand, is initially untrained. The aim of \(G_d\) is to identify the domain of the samples by determining which domain the features extracted by \(G_f\) belong to. The goal of the feature extractor is twofold: first, to extract features that allow \(G_y\) to correctly classify the labels, and second, to fool \(G_d\) so that it cannot correctly predict the domain label. The adversarial relationship between \(G_f\) and \(G_d\) pushes the feature extractor to learn domain-invariant features that make discrimination difficult for \(G_d\). As a result, the extracted features become more robust to domain shifts, effectively aligning the two domains and enabling the model to leverage information from the abundantly labeled source domain to perform well on the sparsely labeled target domain data in the online phase.

The overall loss function is structured as follows:
\begin{equation}
\min_{\theta_f, \theta_y} \max_{\theta_d} \
\Big[
\mathcal{L}_{s}^{y}(\theta_f, \theta_l)
+ \mathcal{L}_{t}^{y}(\theta_f, \theta_l)
-\mathcal{L}^{d}(\theta_f, \theta_d)
\Big]
\end{equation}
where $\theta_f$, $\theta_l$, and $\theta_d$ denote the parameters of the feature extractor $G_f$, label predictor $G_l$, and domain discriminator $G_d$ respectively. There are three loss components, the source domain cross-entropy loss $\mathcal{L}_{s}^{y}$, the target domain cross-entropy loss $\mathcal{L}_{t}^{y}$ which uses $D_t^l$, and the domain discrimination loss $\mathcal{L}^{d}$. The aim of the overall optimization is to minimize the task-related losses $\mathcal{L}_{s}^{y}$ and $\mathcal{L}_{t}^{y}$ with respect to $\theta_f$ and $\theta_l$, ensuring correct label prediction, while simultaneously maximizing the discrimination loss $\mathcal{L}^{d}$.

%% file: 6_results.tex
\section{Results}

\subsection{Dataset Generation}

In our experiments, we have used two widely used, publicly available datasets: RadioML 2016.10a \cite{o2016radio} and RadioML 2018.01a \cite{o2018radio} dataset. These were selected to represent a realistic domain shift to a more complex environment

The Source Domain dataset, used for the offline meta-learning phase, is RML2016.10a. This dataset consists of 220,000 signal examples across 11 modulations. Each input sample is a 256-length vector, representing 128 in-phase and 128 quadrature signal components. The signals span a SNR range from -20 dB to +18 dB in 2 dB increments. The Target Domain dataset, used for the online adaptation phase, is RML2018.01a. It contains over 2.5 million signal examples across 24 modulations. Each input sample is a 2048-length vector, representing 1024 in-phase and 1024 quadrature signal components. The channel simulation is far more challenging, which makes the dataset an ideal target domain to test the model's ability to adapt to more realistic and challenging channel conditions. To construct the datasets for our specific experiment, we first aligned the two datasets by selecting 7 modulation types that exist in both, QAM16, QAM64, AM-DSB, AM-SSB, BPSK, QPSK, 8PSK.

To generate adversarial samples, we created two sets of substitute models. We defined a pool of eleven distinct neural network architectures, namely, VTCNN, VGG16, VGG19, ResNet18, ResNet50, EfficientNet B0, EfficientNet B1, MobileNet V2, MobileNet V3 Small, DenseNet121, and ResNet101. Then, each of these eleven models was trained on the clean RML2016.10a training data to create a set of source substitute models. After that, in a separate process, each model was trained on the RML2018.01a training data to create target substitute models. Adversarial samples for the source domain were then generated by attacking the source substitute models, while target-domain adversarial examples were generated by attacking the target substitute models. For this, we employed a diverse suite of gradient-based and optimization-based attacks, including FGSM \cite{goodfellow2014explainingFSGM}, PGD \cite{mkadry2017towardsPGD,kurakin2016adversarialPGD}, MIM \cite{dong2018boostingMIM}, CW \cite{carlini2017towardsCW}, and PCA \cite{sadeghi2018adversarial}, which we have introduced in Section \ref{preliminaries}. It has to be noted that these models were trained on the datasets for the sole purpose of generating a diverse set of adversarial perturbations and the models themselves are not used as the AMC model.

The combination of our substitute models and attack methods yielded 55 unique pairs for generating distinct adversarial samples. To rigorously test for generalization, our offline phase utilized 50 of these combinations to create meta-training tasks on the source domain data. The final evaluation was then performed on the target domain's held-out test set using the remaining five combinations. This ensures our results reflect the model's performance against attacks that are completely unseen during training, applied to a data distribution that is also new, providing a stringent test of its real-world robustness.

\subsection{Baseline Models}

To evaluate our framework and isolate the contribution of each component, we are using four strategies for the offline phase and three strategies for the online phase to train multiple AMC models, which we introduce here:

Offline Strategies (on Source Domain):
\begin{itemize}
    \item Scratch: (Training from Scratch): No prior offline training.
    \item Transfer Clean: Training on clean data only.
    \item Transfer Adversarial: Standard adversarial training on clean and perturbed data.
    \item Meta Adversarial: Adversarial training using meta-learning across a set of adversarial perturbations. 
\end{itemize}

Online Strategies (on Target Domain):
\begin{itemize}
    \item None (Zero-Shot): Direct evaluation with no adaptation to measure raw generalization.
    \item Fine-tuning: Supervised fine-tuning on a few labeled target samples.
    \item Domain Adaptation: Domain adaptation using DANN on the unlabeled target samples and a few labeled target samples.
\end{itemize}

By combining each offline strategy with each applicable online strategy, we created a comprehensive suite of twelve models for comparison. This approach allows us to systematically measure the performance gains from each component of our proposed framework.

\subsection{Evaluations}

To assess the performance of our proposed framework, we conducted a series of experiments designed to evaluate three key criteria:

\subsubsection{Generalization to Unseen Adversarial Attacks}

\begin{figure}{}
        \centering
        \includegraphics[width=0.8\columnwidth]{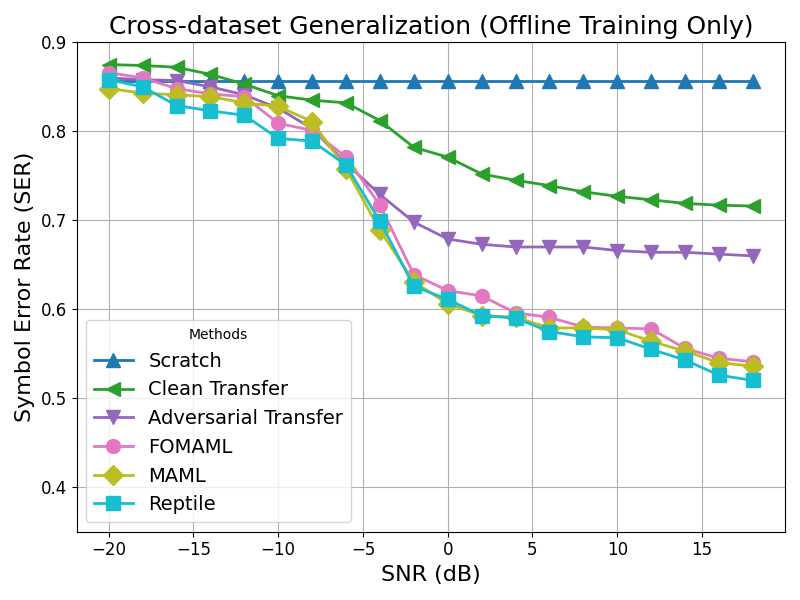} 
        \caption{Cross-dataset Generalization of AMC baselines measured by SER. Models trained on RML2016 during the offline phase, and tested on RML2018 with unseen adversarial attack in the online phase. No fine-tuning during the online phase.}
        \label{fig:generalization}
 \end{figure}

We first aim to to isolate the impact provided by our proposed offline training framework. This is accomplished by comparing the generalization of various AMC models trained on the source domain data during the offline phase and deployed directly on the target domain data without any further fine-tuning, and with previously unseen adversarial attacks. It should be noted that this is a particularly challenging scenario, as adversarial attacks alone can severely degrade AMC performance, and the challenge is compounded by the domain shift between the source and target data with no further trainings. This comparison is illustrated in Figure \ref{fig:generalization}, which visualizes the zero-shot performance across the entire SNR range. As observed, the three methods that follow our proposed adversarial meta-learning offline training framework (Reptile, MAML, and FOMAML) significantly outperform the transfer learning baselines. 

To observe specific numbers, we can look at Table \ref{tab:results}. All methods that have "None" for their Online phase adaptation are relevant to the comparisons in this subsection. As shown, the Scratch model achieves an SER of 0.857. Transfer Learning on clean data reduces this error to 0.796, while Adversarial Transfer Learning further lowers it to 0.66. Our proposed adversarial meta-learning offline training framework achieves the lowest error of 0.53, demonstrating a substantial improvement over all baselines.

\subsubsection{Adaptation Across Domain Shift and Unknown Adversarial Attack}

\begin{figure}[ht]
    \centering

    \subfloat[Standard fine-tuning during the online phase.\label{fig:sub1}]{
        \includegraphics[width=0.7\columnwidth]{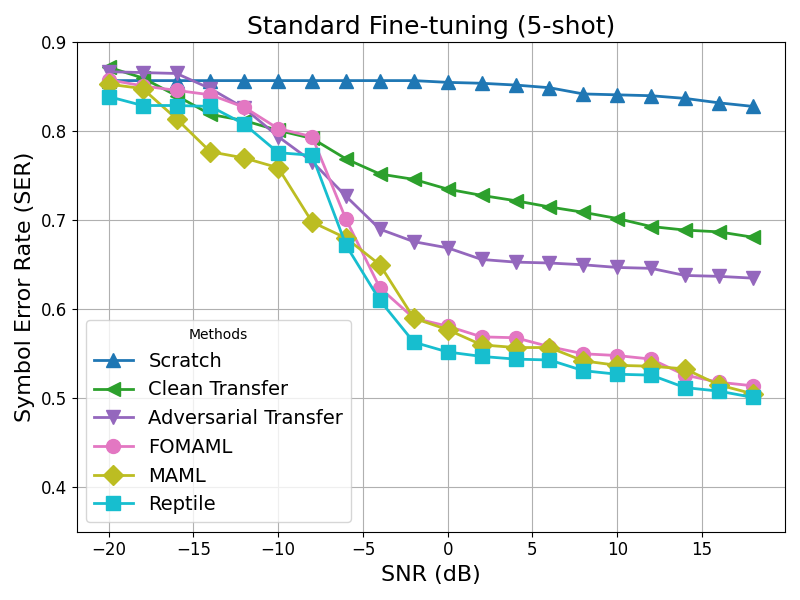}
    }

    \subfloat[Domain adaptation during the online phase.\label{fig:sub2}]{
        \includegraphics[width=0.7\columnwidth]{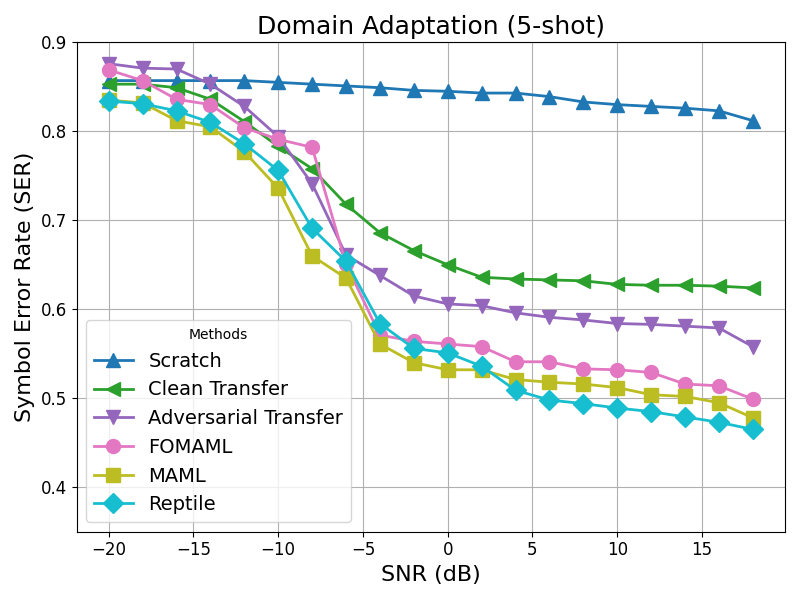}
    }

    \caption{5-shot cross-dataset adaptation of AMC models measured by SER. Models trained on RML2016 during the offline phase, and tested on RML2018 with unseen adversarial attack in the online phase. 5 samples per modulation class from the target dataset, RML2018, available for fine-tuning.}
    \label{fig:5-shot}
\end{figure}

\begin{figure}[ht]
    \centering

    \subfloat[Standard fine-tuning during the online phase.\label{fig:sub1}]{
        \includegraphics[width=0.7\columnwidth]{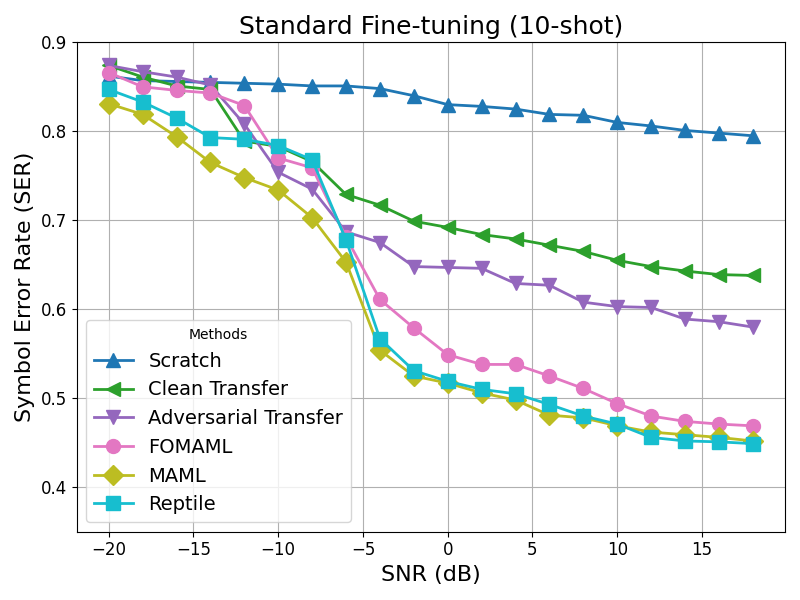}
    }

    \subfloat[Domain adaptation during the online phase.\label{fig:sub2}]{
        \includegraphics[width=0.7\columnwidth]{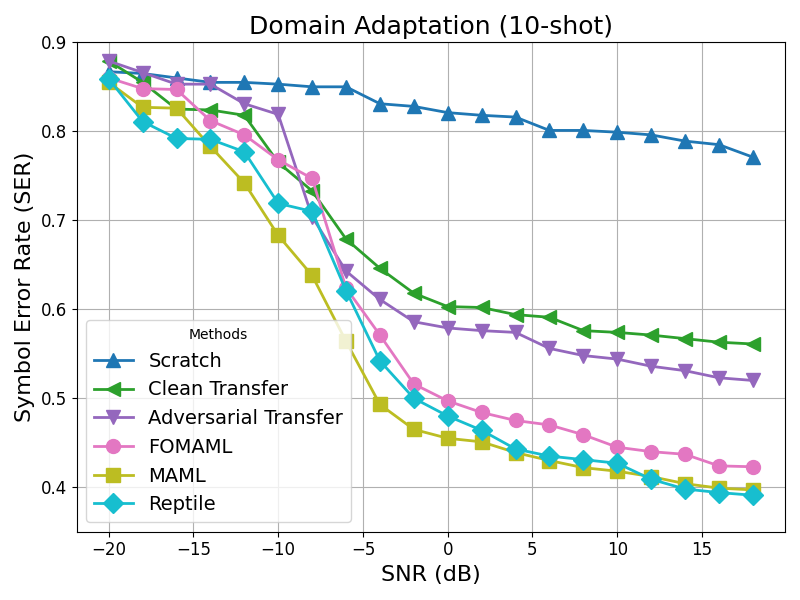}
    }

    \caption{10-shot cross-dataset adaptation of AMC models measured by SER. Models trained on RML2016 during the offline phase, and tested on RML2018 with unseen adversarial attack in the online phase. 10 samples per modulation class from the target dataset, RML2018, available for fine-tuning.}
    \label{fig:10-shot}
\end{figure}

We next evaluate the effectiveness of the proposed online phase domain adaptation framework in transferring the robustness to unseen attakcks in the source domain to the target domain. Figure \ref{fig:5-shot} (a) visualizes the SER of different methods using standard fine-tuning on 5-shot samples (5 labeled samples per class) from the target domain, while Figure \ref{fig:5-shot} (b) shows the SER when the same offline-trained models are fine-tuned using our proposed domain adaptation framework. Comparing the two subfigures, it is evident that the proposed online DA framework consistently reduces SER across all models. Figures \ref{fig:10-shot} (a) and \ref{fig:10-shot} (b) provide the same comparison under 10-shot fine-tuning.

Specific numbers in Table \ref{tab:results} further highlight the improvements. For this subsection, we are interested in looking at rows with "Standard" and "Domain Adaptation" for their Online phase column. For example, our Meta Adversarial model fine-tuned with standard online adaptation achieves a 10-shot SER of 0.452, whereas applying the proposed DA framework reduces it to 0.391. Similarly, Transfer Adversarial with standard adaptation achieves 0.580, while using the DA framework lowers the SER to 0.520. Transfer Clean model benefit from the DA framework, with SER improving from 0.638 to 0.561 under 10-shot fine-tuning. These results demonstrate that our proposed online adaptation framework effectively transfers the robustness learned in the offline phase to the target domain, achieving consistent and significant reductions in error.

\subsubsection{Sample and Computational Efficiency}
Table \ref{tab:efficiency} compares the number of samples required for each model to reach roughly the same level of performance in the online phase, as well as the training time needed to train on that number of samples. Our proposed Meta Adversarial + DA model requires only 5 samples per class for online adaptation, which represents a 3-fold reduction compared to the Transfer Adversarial + DA baseline (15 samples) and a more than 5-fold reduction compared to standard Transfer Clean + DA (26 samples). This is a critical advantage in real-world scenarios where labeled data is often scarce.
This efficiency also extends to computation time. The offline training cost of the proposed Meta Adversarial + DA framework (610.23 s) is higher than that of Transfer Adversarial + DA (419.63 s) and Transfer Clean + DA (382.11 s), while models trained from scratch require no offline training. This additional offline computation enables substantial benefits during online adaptation. The Meta Adversarial + DA model reduces the online adaptation time to 0.464 s compared to 1.749 s for Transfer Adversarial + DA, 2.892 s for Transfer Clean + DA, and 16.803 s for Scratch + DA. Domain adaptation introduces a small increase in online computation compared to standard fine-tuning in the meta-learning setting, but it allows the model to adapt with only 5 samples instead of 8. In the transfer-based models, the reduction in sample requirement is so large that it also translates to lower online computation, with Transfer Clean + DA requiring 26 samples and 2.892 s compared to 57 samples and 3.428 s for Transfer Clean + Standard. The improvement in sample efficiency is highly valuable, making the slight increase in online computation less significant.

\begin{table*}[]
\centering 
\caption{Cross-dataset adaptation of AMC models measured by SER on target dataset. Models trained on RML2016 during the offline phase, and tested on RML2018 with unseen adversarial attack in the online phase. Few (5 and 10) samples per modulation class from the target dataset, RML2018, available for fine-tuning during the online phase.}
\label{tab:results}
\begin{tabular}{|c|c|c|cc|}
\hline
\textbf{Method}                 & \textbf{Offline Phase Training} & \textbf{Online Phase Adaptation} & \multicolumn{1}{c|}{\textbf{5-shot SER}} & \textbf{10-shot SER} \\ \hline
Scratch                         & None                            & None                             & \multicolumn{2}{c|}{0.857}                                      \\ \hline
Scratch + Standard              & None                            & Standard                         & \multicolumn{1}{c|}{0.828}               & 0.795                \\ \hline
Scratch + DA                    & None                            & Domain Adaptation                & \multicolumn{1}{c|}{0.812}               & 0.771                \\ \hline
Transfer Clean                  & Standard                        & None                             & \multicolumn{2}{c|}{0.796}                                      \\ \hline
Transfer Clean + Standard       & Standard                        & Standard                         & \multicolumn{1}{c|}{0.691}               & 0.638                \\ \hline
Transfer Clean + DA             & Standard                        & Domain Adaptation                & \multicolumn{1}{c|}{0.624}               & 0.561                \\ \hline
Transfer Adversarial            & Adversarial                     & None                             & \multicolumn{2}{c|}{0.660}                                      \\ \hline
Transfer Adversarial + Standard & Adversarial                     & Standard                         & \multicolumn{1}{c|}{0.635}               & 0.580                \\ \hline
Transfer Adversarial + DA       & Adversarial                     & Domain Adaptation                & \multicolumn{1}{c|}{0.558}               & 0.520                \\ \hline
Meta Adversarial                & Meta Adversarial                & None                             & \multicolumn{2}{c|}{0.536}                                      \\ \hline
Meta Adversarial + Standard     & Meta Adversarial                & Standard                         & \multicolumn{1}{c|}{0.505}               & 0.452                \\ \hline
Meta Adversarial + DA           & Meta Adversarial                & Domain Adaptation                & \multicolumn{1}{c|}{\textbf{0.465}}      & \textbf{0.391}       \\ \hline
\end{tabular}
\end{table*}

\begin{table*}[]
\centering
\caption{Sample and computation efficiency comparison. Sample and computation needed for different models to approximately reach the same accuracy level in the online phase.}

\label{tab:efficiency}
\begin{tabular}{|c|c|c|c|}
\hline
Method                          & Offline Computation (s) & Online Sample Usage & Online Computation (s) \\ \hline
Scratch + Standard              & \multirow{2}{*}{0}      & 200+                & 12.592                 \\ \cline{1-1} \cline{3-4} 
Scratch + DA                    &                         & 200+                & 16.803                 \\ \hline
Transfer Clean + Standard       & \multirow{2}{*}{382.11} & 57                  & 3.428                  \\ \cline{1-1} \cline{3-4} 
Transfer Clean + DA             &                         & 26                  & 2.892                  \\ \hline
Transfer Adversarial + Standard & \multirow{2}{*}{419.63} & 27                  & 2.127                  \\ \cline{1-1} \cline{3-4} 
Transfer Adversarial + DA       &                         & 15                  & 1.749                  \\ \hline
Meta Adversarial + Standard     & \multirow{2}{*}{610.23} & 8                   & 0.424                  \\ \cline{1-1} \cline{3-4} 
Meta Adversarial + DA           &                         & 5                   & 0.464                  \\ \hline
\end{tabular}
\end{table*}

%% file: 7_conclusion.tex
\section{Conclusion}

The practical deployment of DL-based AMC models is critically hindered by two significant challenges: adversarial attacks and data distribution shifts after deployment during the online phase. In this research, we designed and validated a novel framework that addresses these challenges within a cohesive system through a two-phase learning strategy. The initial offline phase employs meta-learning to build robustness against adversarial perturbations, enabling the model to generalize against unseen black-box attacks. Subsequently, the online phase leverages domain adaptation to maintain classification performance when faced with shifts in the data distribution caused by environmental conditions. Our results demonstrated that this unified framework significantly improves classification accuracy when subjected to these challenging conditions.
The significance of this work lies in bridging the gap between theoretical AMC performance and practical operational reliability. By tackling adversarial vulnerability and domain shift within a single framework, our approach provides a clear path toward developing AMC systems that are not only accurate but also resilient and adaptive. Furthermore, the design principles of our framework can be extended to build more robust models for other wireless tasks, such as RF fingerprinting. This work marks a critical step toward deploying AMC systems that are truly dependable in the unpredictable and contested environments of real-world wireless networks.